\pdfoutput=1

\documentclass[11pt]{article}

\usepackage[final]{acl}
\usepackage[T1]{fontenc} 
\usepackage{arabtex}
\usepackage{utf8}
\setcode{utf8}
\usepackage{times}
\usepackage{latexsym}

\usepackage[utf8]{inputenc}

\usepackage{microtype}

\usepackage{inconsolata}
\usepackage{amsmath}
\usepackage{graphicx}
\usepackage{booktabs}  
\usepackage{array}     
\usepackage{algorithm}
\usepackage{algpseudocode}
\usepackage[section]{placeins}
\usepackage{algorithmicx}

\title{MorphBPE}

\title{MorphBPE: A Morpho-Aware Tokenizer Bridging Linguistic Complexity \\ for Efficient LLM Training Across Morphologies}

\author{
  Ehsaneddin Asgari\textsuperscript{\P,1}, Yassine El Kheir\textsuperscript{*,1,2,3}, Mohammad Ali Sadraei Javaheri\textsuperscript{*,1} \\
    \textsuperscript{1} Qatar Computing Research Institute (QCRI), Doha, Qatar \\
    \textsuperscript{2} German Research Center for Artificial Intelligence (DFKI), Berlin, Germany\\
    \textsuperscript{3} Technical University of Berlin, Berlin, Germany\\
  {\small \textsuperscript{\P} Corresponding Author: easgari@hbku.edu.qa \quad \textsuperscript{*} Equal Contribution}
}

\begin{document}
\maketitle
\begin{abstract}
Tokenization is fundamental to Natural Language Processing (NLP), directly impacting model efficiency and linguistic fidelity. While Byte Pair Encoding (BPE) is widely used in Large Language Models (LLMs), it often disregards morpheme boundaries, leading to suboptimal segmentation—particularly in morphologically rich languages. We introduce MorphBPE, a morphology-aware extension of BPE that integrates linguistic structure into subword tokenization while preserving statistical efficiency. Additionally, we propose two morphology based evaluation metrics: (i) Morphological Consistency F1-Score, which quantifies the consistency between morpheme sharing and token sharing, contributing to LLM convergence, and (ii) Morphological Edit Distance, which measures alignment between morphemes and token concerning interpretability. Experiments on English, Russian, Hungarian, and Arabic across 300M and 1B parameter LLMs demonstrate that MorphBPE consistently reduces cross-entropy loss, accelerates convergence, and improves morphological alignment scores. Fully compatible with existing LLM pipelines, $MorphBPE$ requires minimal modifications for integration. The $MorphBPE$ codebase\footnote{\url{https://github.com/llm-lab-org/MorphBPE}} and tokenizer playground\footnote{\url{https://tokenizer.llm-lab.org}} will be available.\end{list}
\end{abstract}

\setcode{utf8}

\section{Introduction}
Tokenization is a fundamental preprocessing step in NLP, converting raw text into structured units such as bytes~\cite{gillick-etal-2016-multilingual}, characters~\cite{al2019character}, subwords~\cite{sennrich-etal-2016-neural}, words, or multi-word expressions~\cite{gee-etal-2023-multi}. Its effectiveness directly influences downstream tasks, as tokenization errors can propagate through the pipeline, impacting overall model performance \citep{sajjad-etal-2017-challenging, adel2018overview}. Over the years, tokenization has advanced from basic whitespace-based segmentation to sophisticated statistical and neural approaches \citep{smit2014morfessor, otani-etal-2020-pre}. In Large Language Models (LLMs), tokenization significantly affects efficiency, context length, and representational accuracy \citep{dagan2024getting}. Although tokenization-free architectures have been investigated as potential alternatives \citep{clark2022canine, deiseroth-etal-2024-free}, most state-of-the-art models—including Gemma~\cite{team2024gemma}, LLaMA~\cite{touvron2023llama}, DeepSeek~\cite{bi2024deepseek} and OpenAI’s GPT series—still rely on Byte Pair Encoding (BPE)-based tokenization for most languages, retaining both its benefits and inherent limitations.

The additive nature of Byte Pair Encoding (BPE) makes it well-suited for concatenative morphology, as seen in English, where morphemes are linearly appended. However, it struggles with non-concatenative morphological systems, such as root-and-pattern morphology in Arabic and Hebrew, where meaning is encoded through non-linear infixation \cite{khaliq-carroll-2013-induction}. Similarly, agglutinative languages like Turkish, Hungarian, and Korean pose challenges, as their highly productive affixation processes complicate adherence to morpheme boundaries \cite{hakkani-tur-etal-2000-statistical}. These languages require finer-grained tokenization to preserve linguistically meaningful subword structures. Standard BPE and byte-level tokenization methods often struggle to represent these complex morphological patterns effectively, emphasizing the necessity for morphology-sensitive tokenization approaches that better align with the diverse structural properties of different word formation processes \citep{marco-fraser-2024-subword}.

Analyzing BPE output across morphologically rich languages, we observe that its segmentation often disregards meaningful morpheme boundaries, introducing ambiguity and disrupting semantic coherence. For instance, in Arabic, the word \<الرحمن>\ (Al-Rahman, ``The Merciful'') may be incorrectly segmented into \<من>\ (min, ``whom'') \<ال>\ (al, ``the'') + \<رح>\ (rah, an incomplete fragment). Here, \<من>\ (min), a frequent token, is semantically unrelated to the original word, increasing the model's burden in reconstructing meaningful representations. Similar challenges arise in agglutinative and polysynthetic languages, where BPE’s greedy merging strategy often fails to align with true morpheme boundaries.

While purely morphology-based segmentation could mitigate these issues, it has also shown limitations in aligning with naturally occurring linguistic patterns in corpus-based learning \citep{durrani-etal-2019-one,marco-fraser-2024-subword}. Thus, developing tokenization methods that balance morphological integrity with statistical efficiency remains a critical challenge for multilingual NLP.\\

\noindent \textbf{Contributions:} We introduce $MorphBPE$, an extension of Byte Pair Encoding (BPE) that integrates linguistic knowledge into subword tokenization. Our key contributions are:  

\noindent \textbf{(i) A Morphology-Aware LLM Tokenizer:}  
$MorphBPE$ improves adherence to linguistic structures while identifying frequent patterns, balancing \textit{token efficiency} and \textit{interpretability}, particularly in \textit{morphologically rich languages}. It extends BPE by incorporating \textit{morphological structure} while remaining fully compatible with existing LLM training pipelines.  

\noindent \textbf{(ii) Linguistically Informed Tokenizer Evaluation Metrics:}  
We introduce morphology-aware evaluation metrics to assess tokenization quality:  
\begin{itemize}
    \item \textbf{Morph.-Edit Distance Score}: Measures \textit{edit distance} at the \textit{morpheme level}, quantifying segmentation accuracy.  
    \item \textbf{Morph.-Consistency F1-Score}: Inspired by \citep{marco-fraser-2024-subword}, evaluates the \textit{segmentation consistency}, offering a linguistically grounded metric evaluating whether words that share the same morphemes are also assigned the same tokens, and vice versa.
\end{itemize}  
For benchmarking, we curate a dataset covering diverse morphological typologies \cite{ge2022correlations}:  
\begin{itemize}
    \item \textbf{English}: Fusional, low complexity  
    \item \textbf{Russian}: Fusional, moderate complexity  
    \item \textbf{Hungarian}: Agglutinative, high complexity  
    \item \textbf{Arabic}: Templatic, high complexity  
\end{itemize}  
$MorphBPE$ achieves \textit{superior morphological alignment and consistency}, enhancing model interpretability.  

\noindent \textbf{(iii) Empirical Evaluation on LLM Training:}  
We compare $MorphBPE$ to vanilla BPE on \textit{300M} and \textit{1B} parameter LLMs across the four languages, demonstrating:  
\begin{itemize}
    \item \textbf{Lower training loss}, indicating improved linguistic representations.  
    \item \textbf{Faster convergence}, enhancing computational efficiency.  
\end{itemize}  

By integrating linguistic principles with modern tokenization strategies, MorphBPE bridges the gap between traditional morphological analysis and NLP, providing a computationally efficient and morphologically interpretable tokenization approach for language modeling, particularly in morphologically rich languages like Arabic. In line with this, MorphBPE\footnote{MorphBPE is provisionally patented under U.S. Provisional Patent Application No. 63/679,403.} has been developed and implemented in \textbf{Fanar}\footnote{www.fanar.qa}, an Arabic-centric language model, leading to significant improvements in model performance~\cite{team2025fanar}.

\section{Related Work}
BPE, originally introduced as a text compression algorithm \citep{shibata1999byte}, was first adapted for machine translation as a tokenization method in 2016 \citep{sennrich-etal-2016-neural}. Since then, it has become the de facto standard in NLP and Large Language Models (LLMs) due to its efficiency in managing vocabulary size, handling out-of-vocabulary words, and capturing frequent patterns, while offering partial improvements over morphology-based tokenizers \citep{sennrich-etal-2016-neural}.  

Despite its widespread adoption, vanilla BPE has several notable limitations: its greedy merging strategy, inefficiencies in cross-lingual settings where similar words with different character variations are not aligned, and inconsistent handling of character-level information across languages. To address these challenges, various extensions have been proposed, including BPE dropout \citep{provilkov-etal-2020-bpe}, which introduces stochasticity to improve generalization, sampling-based BPE \citep{asgari2019probabilistic, asgari2020subword}, which enhances subword diversity, byte-level adaptations \citep{wang2020neural}, which aim to improve robustness across scripts, and multilingual BPE variants \citep{liang-etal-2023-xlm}, designed to optimize token sharing across languages.  

The importance of morphology-aware tokenization for language models has been recognized in several recent studies \citep{park-etal-2021-morphology, jabbar2023morphpiece, marco-fraser-2024-subword, weller-di-marco-fraser-2024-analyzing}. However, an integrated solution that effectively balances morphological information with frequent pattern extraction while remaining fully compatible with modern LLM training pipelines has remained an open problem.  
 \section{Methods}

\begin{figure*}[ht]
    \centering
    \resizebox{2\columnwidth}{!}{
        \includegraphics[trim=0cm 0cm 0cm 0cm, clip]{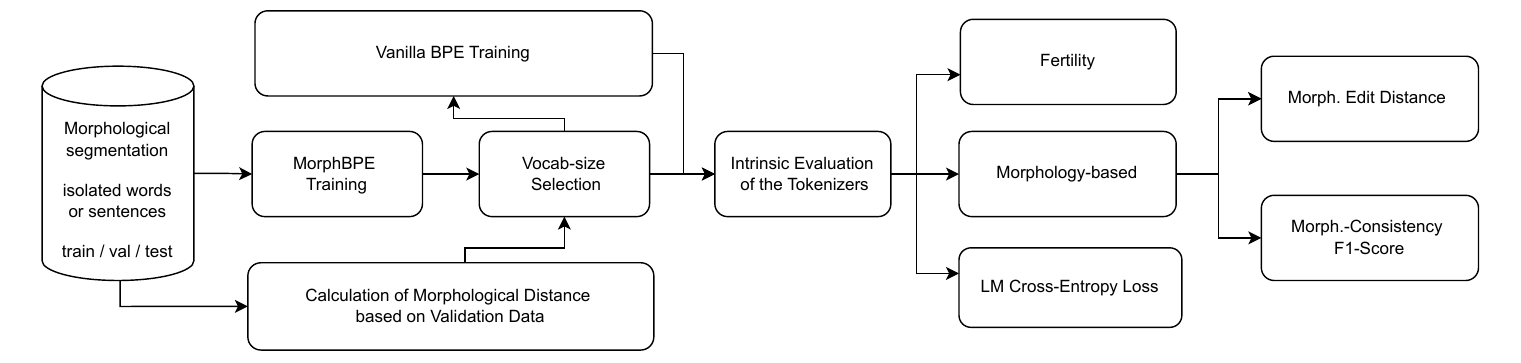}
    }
    \caption{Overview of the $MorphBPE$ study: We evaluate the effectiveness of $MorphBPE$ over vanilla BPE across four morphologically diverse languages (English, Russian, Hungarian, and Arabic) by aligning vocabulary size with morphological segmentation. The we evaluate the tokenizers using the intrinsic evaluation metrics.}
    \label{fig:morphbpe_overview}
\end{figure*}

Figure~\ref{fig:morphbpe_overview}
 provides an overview of our approach. To systematically evaluate \textbf{MorphBPE}, we select four languages with distinct morphological typologies, where morphological segmentation is available for training and evaluation at the word level. We determine the vocabulary sizes for each language based on optimal alignment with morphological boundaries. Then we evaluate the vanilla BPE and $MorphBPE$ on the selected vocabulary size using intrinsic metrics detailed in \S\ref{sec:metrics}.

\subsection{Datasets}

\subsubsection{Morphological Data}
Our dataset comprises morphologically segmented words from four morphologically diverse languages \cite{ge2022correlations}: English, Russian, Hungarian, and Arabic. The segmentation data for English, Russian, and Hungarian is sourced from the \textit{SIGMORPHON 2022 Shared Task on Morpheme Segmentation} \cite{batsuren-etal-2022-sigmorphon}, which provides high-quality morpheme segmentations. To incorporate a \textit{root-based (templatic)} morphological system, we include Arabic, where, we utilize multiple sources: the \textit{Arabic Treebank (ATB)} dataset \cite{taji-etal-2017-universal}, the \textit{Dialectal Segmentation Dataset} \cite{darwish-etal-2018-multi}, and \textit{Quranic morphology data} \cite{dukes-habash-2010-morphological}. Additionally, we enrich this set with $1M$ high-confidence segmentations of frequent Arabic surfaceforms obtained using \textit{Farasa} \cite{darwish-mubarak-2016-farasa}. All datasets were cleaned and standardized. Manually annotated segmentations were split into $80\%$ training, $10\%$ validation, and $10\%$ test sets. Table~\ref{tab:morph_segmentation} summarizes the dataset composition.

\begin{table*}
\centering
\caption{Token Statistics for Morphological Segmentation Datasets Used in BPE and $MorphBPE$ Training and Tokenizer Evaluation Across Languages.}
\label{tab:morph_segmentation}
\resizebox{1.6\columnwidth}{!}{%
\begin{tabular}{llcc}
\toprule
 Language & Morphology Type & \# of Words & Avg. Morphemes per Word \\
\midrule
Hungarian & Agglutinative & 930,312 & 3.22 \\
Russian & Fusional (moderate complexity) & 784,212 & 3.84 \\
English & Fusional (low complexity) & 571,495 & 2.33 \\
Arabic & Root-based (Templatic) & 1,395,835 & 2.50 \\
\bottomrule
\end{tabular}}
\end{table*}

\begin{table*}[h]
\centering
\caption{Morph.-consistency evaluation: Precision, Recall, and F1-score for BPE and $MorphBPE$ in different languages. A higher F1-score ($\mu_c$) indicates greater consistency in segmenting words with similar or dissimilar morphemes. Results are reported as mean $\pm$ standard deviation over multiple resamples over test sets.}
\label{tab:morphconsistency}
\resizebox{2\columnwidth}{!}{\begin{tabular}{l|cc|c}
\hline
\textbf{Model} & \textbf{Precision (Mean ± Std)} & \textbf{Recall (Mean ± Std)} & \textbf{Morph.-Consistency F1-score ($\mu_c$)} \\
\hline
English BPE (96K) & 0.00 ± 0.00 & 0.03 ± 0.02 & 0.00 \\
English $MorphBPE$ (96K) & 0.20 ± 0.42 & 0.30 ± 0.06 & \textbf{0.24} \\\hline

Russian BPE (64K) & 0.10 ± 0.32 & 0.06 ± 0.01 & 0.07 \\
Russian $MorphBPE$ (64K) & 0.69 ± 0.48 & 0.33 ± 0.06 & \textbf{0.45} \\\hline

Hungarian BPE (24K) & 0.08 ± 0.25 & 0.29 ± 0.04 & 0.13 \\
Hungarian $MorphBPE$ (24K) & 0.98 ± 0.03 & 0.78 ± 0.07 & \textbf{0.87} \\\hline

Arabic BPE (96K) & 0.00 ± 0.00 & 0.08 ± 0.03 & 0.00 \\
Arabic $MorphBPE$ (96K) & 0.89 ± 0.31 & 0.53 ± 0.05 & \textbf{0.66} \\

\hline
\end{tabular}}
\end{table*}

\begin{figure}
    \centering
    \includegraphics[width=1\columnwidth]{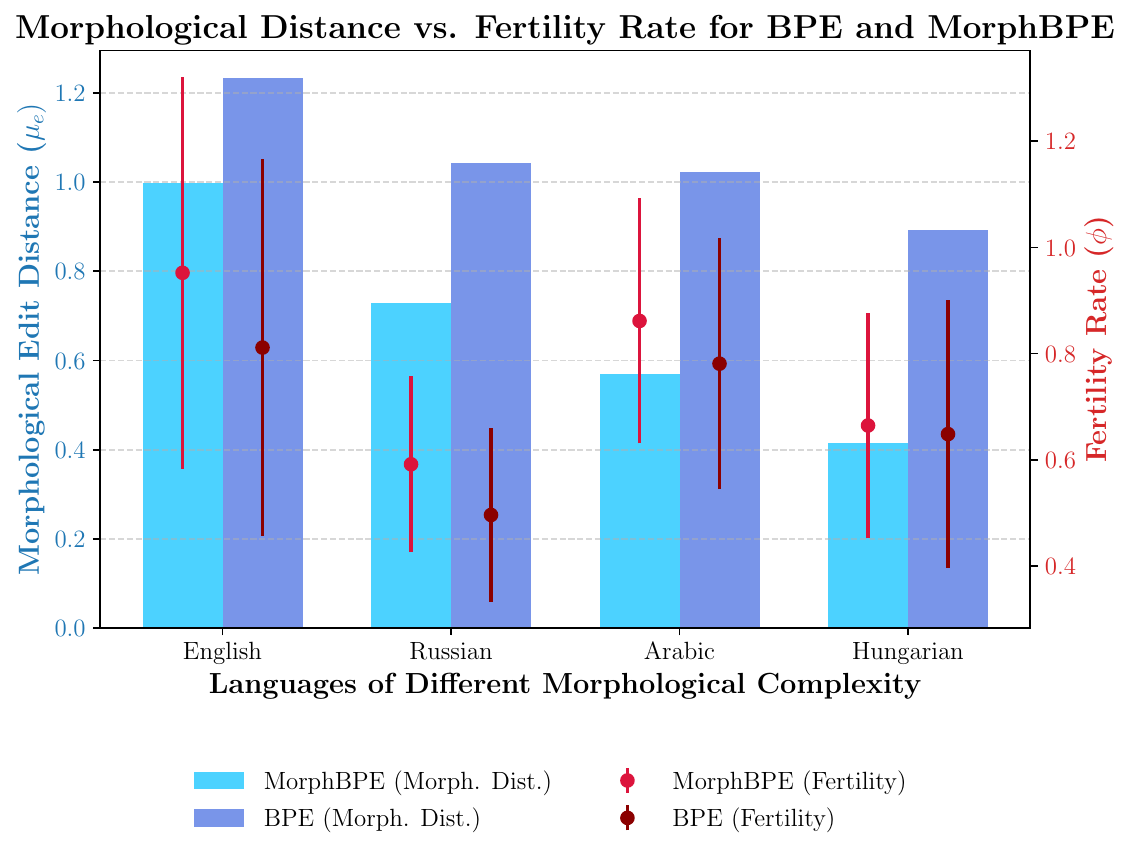}
    \caption{Comparison of morphological distance and fertility rate for BPE and $MorphBPE$ across four languages.}
    \label{fig:morph_fertility}
\end{figure}

\subsubsection{LLM Training Data}
For our study on Evaluating $MorphBPE$ vs. BPE Across Languages with Diverse Morphologies: Hungarian, Arabic, Russian, and English, we require a large-scale multilingual training dataset. We selected \textbf{FineWeb2} \cite{penedo2024fineweb-2}, a comprehensive corpus covering over 1,000 languages, to ensure sufficient tokens for training, following the \textit{Chinchilla scaling law} \cite{10.48550/arxiv.2203.15556}. This choice enables a balanced token distribution across the selected languages, ensuring fair and robust evaluation of $MorphBPE$ and BPE. 

\subsection{MorphBPE approach}
\textit{MorphBPE} is a simple yet effective extension of BPE that prevents frequent symbol pair merges from crossing morpheme boundaries while keeping the rest of the algorithm unchanged (Algorithm~\ref{alg:morphbpe}). This ensures compatibility with standard BPE inference, making $MorphBPE$ easy to integrate into existing pipelines without modifications. 

\begin{algorithm}\small
\caption{\small{Morphologically-aware Byte Pair Encoding (MorphBPE)}}
\label{alg:morphbpe}
\begin{algorithmic}[1]
\State Initialize vocabulary with individual characters
\State Segment the training corpus using morphological segmentation
\While{number of merges $<$ desired vocabulary size}
    \State Compute byte-pair frequencies
    \State \textbf{Morph-aware Step:} Merge the most frequent byte pair without crossing morpheme boundaries
    \State Update vocabulary with the merged symbol
\EndWhile
\end{algorithmic}
\end{algorithm}

\subsection{Tokenization Evaluation}
\label{sec:metrics}
Tokenization evaluation can be conducted using intrinsic or extrinsic metrics. Extrinsic evaluation assesses tokenizers in the broader context of LLM performance across diverse capabilities, requiring extensive pre/post training and high-level analysis, which is beyond the scope of this work~\cite{cecchini-etal-2024-holistic,chia-etal-2024-instructeval}. Before evaluating tokenizers in downstream tasks, it is essential to first examine fundamental properties to ensure efficiency and consistency. Therefore, we focus on intrinsic evaluation metrics that provide insights into the core characteristics of tokenization in large language models (LLMs).

\noindent \textbf{(i) Fertility ($\phi$):} Fertility quantifies the number of tokens generated by a tokenizer relative to a baseline, typically a whitespace-based tokenizer~\cite{rust-etal-2021-good}. A lower fertility score generally indicates a more efficient representation, enabling longer contexts. However, this assumption is debatable, particularly for agglutinative languages such as Hungarian and Turkish, where capturing morphological structure necessitates more tokens to provide adequate context for each surface form. As shown in Table~\ref{tab:morph_segmentation}, languages vary in the average number of morphemes per word. For instance, Hungarian and Arabic require more tokenization compared to English to accurately represent their linguistic structures.

\noindent \textbf{(ii) Morph.-Edit Distance Score ($\mu_e$):} We introduce a new intrinsic evaluation metric, the \textit{morphological edit distance}, which assesses how well tokenization aligns with the underlying morphological segmentation of words. This metric is computed using a pairwise alignment score based on dynamic programming, ensuring that the order of matching tokens with segmented morphemes is preserved. This approach quantitatively evaluates how effectively a tokenizer respects the morphological structure of the language. We refer to this metric as the Morphology Edit Distance Score ($\mu_e$), which evaluates the interpretability of the tokenizer. While it can be normalized by the number of morphemes in each word, we retain its raw form to provide a clearer indication of the average number of edits required.
\begin{figure*}[ht!]
    \centering
    \includegraphics[width=1.1\textwidth, trim=2cm 0 0 0, clip]{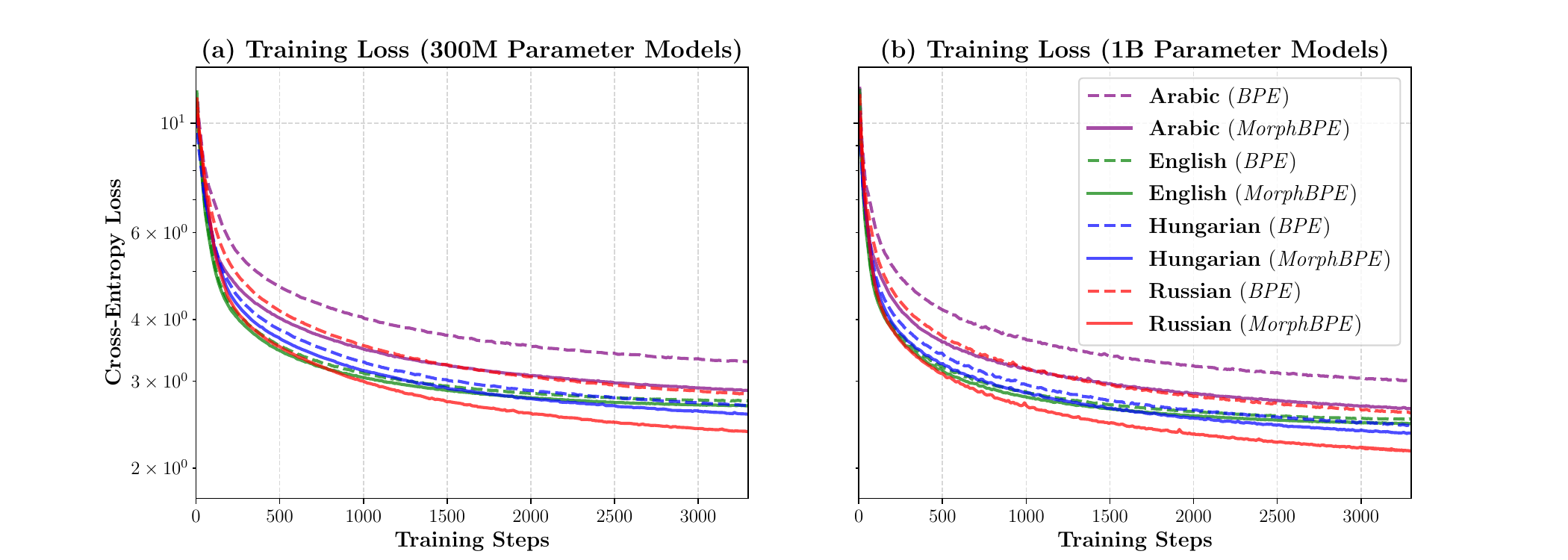}
    \caption{Comparison of training cross-entropy loss between BPE and $MorphBPE$ across four languages. Results are shown for both the small (300M) and large (1B) models.}
    \label{fig:cross_entropy_loss}
\end{figure*}
\noindent \textbf{(iii) Morph.-Consistency Scores (F1: $\mu_c$):}  Inspired by the discussion in \citep{marco-fraser-2024-subword}, we propose a morphology consistency measure, which is crucial for language model training. It ensures that words sharing the same morphemes also share tokens (recall score) and that words with shared tokens correspondingly share morphemes (precision score). This evaluation is conducted over a dataset of segmented words, where shared morpheme/token relationships can be treated as either binary events or weighted counts. For simplicity, we adopt a binary scheme, checking whether shared morphemes correspond to shared tokens and vice versa. Since both precision and recall are essential for avoiding unnecessary ambiguity and maintaining a consistent representation of related words, we use their harmonic mean, i.e., the F1-score of morphological consistency, denoted as $\mu_c$. 

To ensure practical feasibility given large evaluation datasets, we employ $k$-means clustering ($k=100$) to group words with similar morphemes and measure scores between $C=50$ word pairs within each cluster. Precision and recall are estimated through a bootstrapping procedure, drawing $N=10$ resamples from clusters.

\subsection{Vocabulary Size Selection}
Vocabulary size is a critical hyperparameter in LLM training, directly impacting model performance across languages. To determine the optimal vocabulary size in $MorphBPE$, for our four languages, we employed a morphology distance score, $\mu_e$, computed over the development set. We evaluated vocabulary sizes from 8K to 96K in 8K increments, selecting the smallest size beyond which further increases did not yield statistically significant improvements in morphological alignment (\textit{measured via a t-test over the dev. vocabularies}). Through this approach, we determined optimal sizes of \textbf{24K for Hungarian} and \textbf{64K for Russian}, where larger vocabularies showed diminishing returns. For \textbf{English and Arabic}, morphology distance continued improving with larger vocabularies, leading us to select \textbf{96K}. 

We evaluated the selected tokenizers based on (i) fertility rate ($\phi$), (ii) morphological edit distance score ($\mu_e$), and (iii) morphological consistency score ($\mu_c$) on the test sets of English, Russian, Hungarian, and Arabic. Since fertility rate is a relative measure, we compare both $MorphBPE$ and BPE against a strong multilingual baseline—Bloomz (256K) \cite{yong-etal-2023-bloom}, which employs a large vocabulary to accommodate multiple languages. In contrast, $\mu_e$ and $\mu_c$ are directly computed from the test data to evaluate tokenization quality with respect to linguistic structure.

\noindent \textbf{(iv) Cross Entropy Loss of Language Modeling ($l_c$)}  
Cross-entropy loss in language modeling measures the divergence between predicted and ground truth outputs. The trajectory of training cross-entropy loss indicates how quickly a model converges and improves next-token prediction. This metric is closely related to model perplexity, a standard intrinsic evaluation measure for language models. However, cross-entropy loss is only comparable across models with identical vocabulary sizes, as vocabulary variations directly affect the model's branching factor.

\subsection{Language Model Training}
To assess the scalability of our approach, we trained two model sizes—\textbf{300M (small)} and \textbf{1B (large)}—using decoder architectures within the LLaMA-Factory framework \cite{zheng-etal-2024-llamafactory}. For each language, we trained models with both vanilla BPE and $MorphBPE$ of the same vocabulary sizes, resulting in four models per language. Training loss was monitored and compared across languages and tokenization methods to evaluate their impact on learning efficiency. We ensured passing $\approx6B$ tokens to the small and $\approx20B$ tokens to the large model compatible with the \textit{Chinchilla scaling law} \cite{10.48550/arxiv.2203.15556}.

\section{Results}

\subsection{Morphological Metrics and Fertility}
The results in Figure~\ref{fig:morph_fertility} and Table~\ref{tab:morphconsistency} show a clear trend: \textit{MorphBPE} consistently achieves lower morphological edit distance ($\mu_e$) and higher morphological consistency ($\mu_c$) compared to BPE, with a slight increase in fertility rate across all languages. The extent of improvement varies based on the morphological complexity of the language. The gap in $\mu_e$ and $\mu_c$ between \textit{MorphBPE} and BPE is larger for Hungarian and Arabic, which have more complex morphological structures. These results indicate that \textit{MorphBPE} better preserves linguistic structure, particularly in morphologically rich languages, while BPE tends to over-fragment words based on subword frequency rather than morpheme boundaries. Higher $\mu_c$ of \textit{MorphBPE} also reflects consistent tokenization which morphology, which can impact the convergence of language model training.

\subsection{Training cross-entropy loss}

The training cross-entropy loss for the four languages, using the same vocabulary and comparing BPE and $MorphBPE$, is presented in Figure~\ref{fig:cross_entropy_loss}. The results are shown over a training window of $\approx14B$ tokens for both small and large models, with the selected interval chosen for clarity, as the overall trend remains consistent throughout training. The results indicate that $MorphBPE$ consistently improves cross-entropy loss across all languages and model sizes, even for English language. This improvement is particularly pronounced in morphologically richer languages, where the reduction in loss is more significant. The results demonstrate lower training loss, indicating improved linguistic representations as well as faster convergence. 


\section{Discussions and Conclusion}

In this work, we introduced $MorphBPE$, a morphology-aware extension of BPE that integrates linguistic knowledge into subword tokenization. Through extensive empirical evaluation across English, Russian, Hungarian, and Arabic, we demonstrated that $MorphBPE$ consistently enhances LLM training efficiency by reducing cross-entropy loss, improving morphological alignment, and accelerating convergence across both 300M and 1B parameter models.

Another key contribution of this work is the introduction of linguistically informed tokenizer evaluation metrics, addressing a critical gap in current tokenization evaluation. The Morphological Consistency F1-Score provides a structured measure of segmentation stability, which is essential for ensuring consistent morpheme-level representations during LLM training. This stability directly contributes to better generalization and improved learning efficiency, particularly for morphologically rich languages. Meanwhile, the Morphological Alignment Score, based on edit distance at the morpheme level, serves as a linguistically grounded metric, that can contribute to the interpretability of the tokenizer.

We show that $MorphBPE$, despite having higher fertility, results in a more interpretable and more consistent and more efficient tokenizer for LLM training. This suggests that fertility—a commonly used metric in tokenization evaluation—may not be the most reliable indicator of tokenizer quality of an efficient LLM training. 

An additional advantage of $MorphBPE$ is its full compatibility with existing LLM training and inference pipelines, requiring minimal modifications to the tokenization process. This ensures easy integration without disrupting standard workflows. Furthermore, an efficient implementation of $MorphBPE$ training and evaluation metrics will be released with this work, enabling reproducibility and facilitating further research in morphology-aware tokenization.

\section{Limitations}
We demonstrated the effectiveness of $MorphBPE$ across four languages with diverse morphological typologies. However, future work can extend this evaluation to additional languages. One limitation is that $MorphBPE$ relies on the availability of morphological segmentation data, which is not yet accessible for all languages. Efforts such as UniMorph~\cite{kirov-etal-2018-unimorph} and MorphyNet~\cite{batsuren-etal-2021-morphynet} are helping bridge this gap, but further development is needed. Additionally, an important next step is the extrinsic evaluation of LLMs trained with $MorphBPE$, assessing their impact on higher-level capabilities.

\section*{Acknowledgments}

We extend our gratitude to Sanjay Chawla, Mohamed Eltabakh, Ahmed Ali, Muhammad Tasnim Mohiuddin, Sabri Boughorbel, Hamdy S. Mubarak, MohammadAmin Sadeghi, Natasa Milic-Frayling, Ali Nazari, Nadir Durrani, Mohsen Mahdavi Mazdeh, and the entire Fanar team for their valuable feedback and insights.


\bibliography{anthology,custom}




\end{document}